\documentclass[a4paper,twoside]{article}

\usepackage{epsfig}
\usepackage{multicol}
\usepackage{pslatex}
\usepackage{natbib}
\usepackage{wrapfig}
\usepackage{multirow}
\usepackage[utf8]{inputenc} 
\usepackage[T1]{fontenc}    
\usepackage{url}            
\usepackage{booktabs}       
\usepackage{amsfonts}       
\usepackage{nicefrac}       
\usepackage{microtype}      

\usepackage[]{hyperref}
\hypersetup{
    colorlinks=true,
    allcolors=CetaceanBlue,
    linkcolor=CetaceanBlue,
    filecolor=CetaceanBlue,
    urlcolor=CetaceanBlue,
}

\usepackage{csquotes}
\usepackage{dirtree}

\usepackage{xcolor}
\definecolor{CetaceanBlue}{RGB}{0 21 64}

\usepackage{graphicx}
\graphicspath{ {images/} }

\newcommand{\tabularspace}{\vspace{3pt}}

\usepackage{cleveref}

\usepackage{SCITEPRESS}     

\begin{document}

\title{The MVTec 3D-AD Dataset for \\ Unsupervised 3D Anomaly Detection and Localization}

\author{
\authorname{%
Paul Bergmann\sup{1,2}\orcidAuthor{0000-0002-4458-3573},
Xin Jin\sup{3}\orcidAuthor{0000-0002-2078-8056}, 
David Sattlegger\sup{1}\orcidAuthor{0000-0002-8336-4672} and 
Carsten Steger\sup{1}\orcidAuthor{0000-0003-3426-1703}}
\affiliation{\sup{1}MVTec Software GmbH, Germany}
\affiliation{\sup{2}Technical University of Munich, Germany}
\affiliation{\sup{3}Karsruhe Institute of Technology, Germany}
\email{paul.bergmann@mvtec.com, xinjin.kit@gmail.com, sattlegger@mvtec.com, steger@mvtec.com}
}


\abstract{We introduce the first comprehensive 3D dataset for the task of unsupervised anomaly detection and localization. It is inspired by real-world visual inspection scenarios in which a model has to detect various types of defects on manufactured products, even if it is trained only on anomaly-free data. There are defects that manifest themselves as anomalies in the geometric structure of an object. These cause significant deviations in a 3D representation of the data. %
We employed a high-resolution industrial 3D sensor to acquire depth scans of 10 different object categories. For all object categories, we present a training and validation set, each of which solely consists of scans of anomaly-free samples. The corresponding test sets contain samples showing various defects such as scratches, dents, holes, contaminations, or deformations. Precise ground-truth annotations are provided for every anomalous test sample. %
An initial benchmark of 3D anomaly detection methods on our dataset indicates a considerable room for improvement.
}

\onecolumn \maketitle \normalsize \setcounter{footnote}{0} \vfill

\section{\uppercase{Introduction}}
The increased availability and precision of modern 3D sensors has led to significant advances in the field of 3D computer vision. The research community has used these devices to create datasets for a wide variety of real-world problems, such as point cloud registration \citep{Zeng20163DMatch}, classification \citep{Wu2015Modelnet40}, 3D semantic segmentation \citep{Chang2015Shapenet,Dai2017Scannet}, 3D object detection \citep{Armeni20163DSemanticParsing}, and rigid pose estimation \citep{Drost2017Itodd,Hodan2020Bop}. The development of new and improved algorithms relies on the availability of such high quality datasets.

\begin{figure*}[ht]
    \centering
    \includegraphics[width=\textwidth]{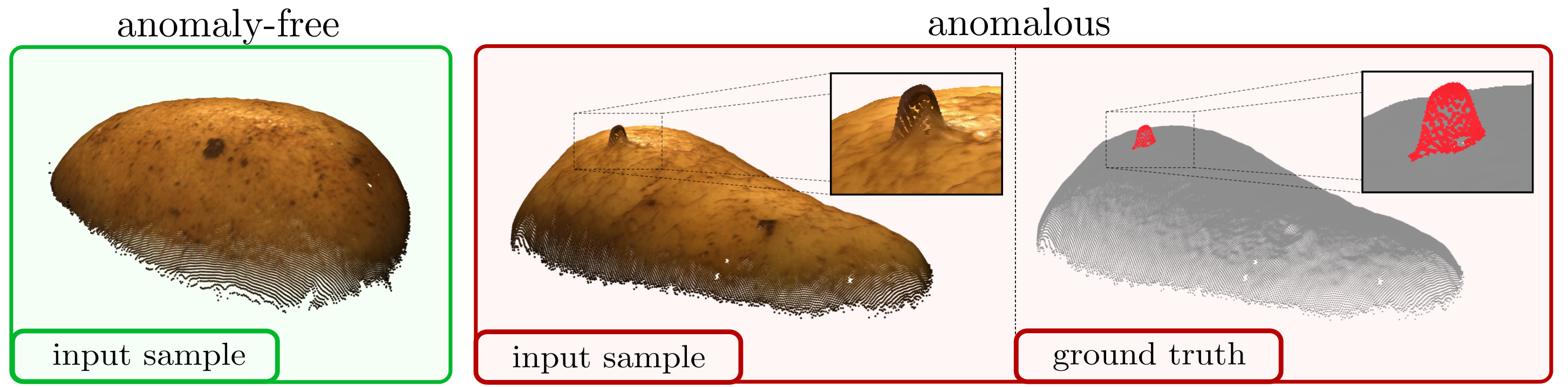}
    \caption{Two samples representing the category \textit{potato} of our new dataset. The sample on the left is anomaly-free while the one on the right contains a nail stuck through the surface of the potato. We depict the point clouds overlaid with the associated color values. For the anomalous sample, we additionally display the annotated ground truth.}
    \label{fig:teaser}
\end{figure*}

A task of particular importance in many applications is to recognize anomalous data that deviates from what a model has previously observed during training. In manufacturing, for example, these kinds of methods can be used to detect defects during inference, while only being trained on anomaly-free samples. In autonomous driving, it is safety critical that an intelligent system can detect structures it has not seen during training. This problem has attracted significant attention for color or grayscale images. Curiously, the field of unsupervised anomaly detection is comparatively unexplored in the 3D domain. We believe that a key reason for this is the unavailability of suitable datasets.
 
To fill this gap and spark further interest in the development of new methods, we introduce a real-world dataset for the task of unsupervised 3D anomaly detection and localization. Given a set of exclusively anomaly-free 3D scans of an object, the task is to detect and localize various types of anomalies. Our dataset is inspired by industrial inspection scenarios. This application has been identified as an important use case for the task of unsupervised anomaly detection because the nature of all possible defects that may occur in practice is often unknown. In addition, defective samples for training can be difficult to acquire and precise labeling of defects is a laborious task. \Cref{fig:teaser} shows two prototypical samples of our new dataset.
Our main contributions are as follows:
\begin{itemize}
    \item We introduce the first comprehensive dataset for unsupervised anomaly detection and localization in three-dimensional data. It consists of 4147 high-resolution 3D point cloud scans from 10 real-world object categories. While the training and validation sets only contain anomaly-free data, the samples in the test set contain various types of anomalies. Precise ground truth annotations are provided for each anomaly.\footnote{The dataset will be made publicly available at \href{https://www.mvtec.com/company/research/datasets}{https://www.mvtec.com/company/research/datasets}.}

    \item We evaluate current methods that were specifically designed for unsupervised 3D anomaly localization. Our initial benchmark reveals that existing methods do not perform well on our dataset and that there is considerable room for future improvement.
\end{itemize}


\section{\uppercase{Related Work}}

\paragraph{Anomaly Detection in 2D.} 

For two-dimensional image data, numerous synthetic and real-world benchmark datasets exist. They cover various domains, e.g., autonomous driving \citep{fishyscapes2019}, video anomaly detection \citep{ucsd_video_ad_dataset, avenue_video_ad_dataset,Sultani_2018_CVPR_videoAD_dataset}, or industrial inspection scenarios.

Since we propose a new industrial inspection dataset, we focus our summary on existing datasets from this domain. The task is to detect and localize defects on manufactured products when only a set of anomaly-free training images is available. \citet{Huang2018MagneticTileDataset} present a surface inspection dataset of 1344 images of magnetic tiles. Test images contain various types of anomalies, such as cracks or uneven areas. Similar datasets exist that focus on the inspection of a single repetitive two-dimensional texture \citep{Carrera2017HandcraftedFeatureDictionaryNanofibres, Song2013SteelSurfaceDefectDatabase, Wieler2007DAGMDataset}. \citet{bergmann2019mvtec,Bergmann2021IJCV} introduce a more comprehensive dataset that contains a total of 5354 images showing instances of five texture and ten object categories. The test set contains 73 distinct types of anomalies, such as contaminations or scratches on the manufactured products. 

The aforementioned datasets have led to the development of numerous methods that are intended to operate on 2D color images. A popular approach \citep{Bergmann_2020CVPR, cohen2021subimage, Salehi2021CVPR, Wang2021CVPR} is to model the distribution of descriptors extracted from neural networks pretrained on large-scale datasets like ImageNet \citep{krizhevsky2012imagenet}. Networks that are pretrained in such a way expect RGB images as input. The resulting methods are therefore ill-suited to process 2D representations of 3D data, such as depth images, and cannot be easily transferred to 3D anomaly detection. A different line of work uses generative models such as convolutional autoencoders (AEs) \citep{Masci2011CAE} or generative adversarial networks (GANs) \citep{goodfellow_GAN} to detect anomalies by evaluating a pixelwise reconstruction error. \citet{fast_anogan_schlegl} introduce f-AnoGAN, where a GAN is trained on the anomaly-free training data. In a second step, an encoder network is trained to output latent samples that reconstruct the respective input images when passed to the generator of the GAN. Similarly, methods based on autoencoders \citep{bergmannp2019ssim,Park2020MNAD} first encode input images with a low dimensional latent sample, and then decode that sample to minimize a pixelwise reconstruction error. For both approaches, anomaly scores are computed by a pixelwise comparison of an input image with its reconstruction. Since these methods do not require a domain-specific pretraining, they can be adapted to other two-dimensional representations, such as depth images. 

\begin{figure*}[t]
  \centering
  \includegraphics[width=0.91\textwidth]{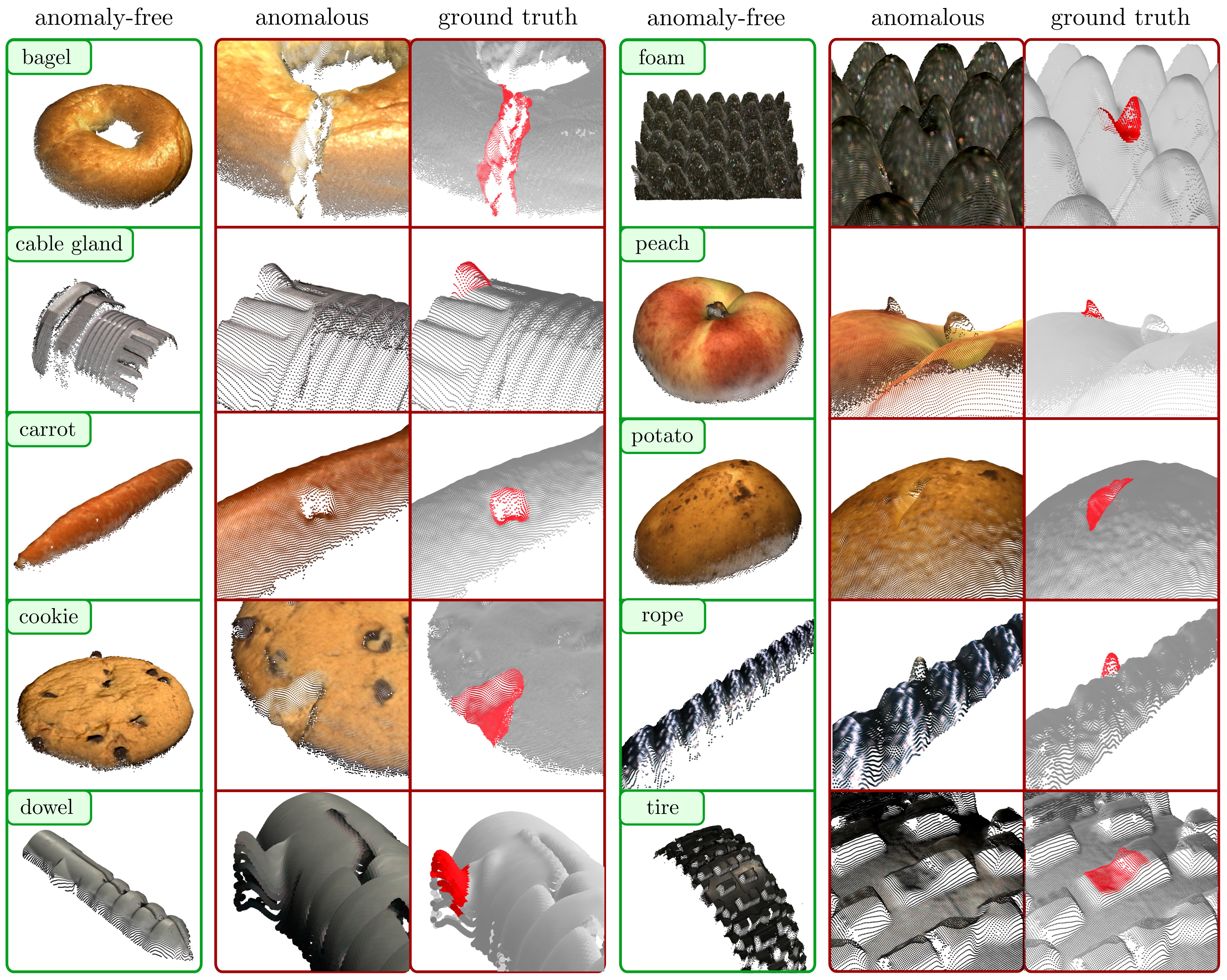}
  \caption{Examples for all 10 dataset categories of the MVTec 3D-AD dataset. For each category, the left column shows an anomaly-free point cloud with RGB values projected onto it. The second column shows a close-up view of an anomalous test sample. Anomalous points are highlighted in the third column in red. Note that the background planes were removed for better visibility.}
  \label{fig:dataset_overview}
\end{figure*}

\paragraph{Anomaly Detection in 3D.} 

To date, there exists no comprehensive 3D dataset designed explicitly for the unsupervised detection and localization of anomalies. Existing methods for this problem are evaluated on two medical benchmarks that were originally introduced for the supervised detection of diseases in brain magnetic resonance (MR) scans. \citet{brats2015}, \citet{Bakas_2017}, and \citet{brats2021} present the multimodal brain tumor image segmentation benchmark (BRATS). It consists of 65 multi-contrast MR scans from glioma patients. Each sample is provided as a dense voxel grid and tumors were annotated by radiologists in each image slice of the scan. Similarly, \citet{PMID:29461514} provide the Anatomical Tracings of Lesions After Stroke (ATLAS) dataset. It consists of 304 MR scans with corresponding ground truth annotations of brain lesions. Both these datasets provide 3D information by stacking multiple grayscale images to form a dense voxel grid. Hence, the nature of this data is fundamentally different from the one in our dataset that describes the geometric surface of objects.

Due to the lack of more diverse 3D datasets for unsupervised anomaly detection, there exist only very few methods designed for this task. Only recently, \citet{Simarro_Viana_2021} introduced an extension of f-AnoGAN to voxel data. As for the 2D case, a GAN is trained to generate voxel grids that mimic the training distribution using 3D convolutions. Subsequently, an encoder network is trained that maps input samples to the corresponding latent samples of the generator. During inference, anomaly scores are derived for each voxel element by comparing the input voxel to the reconstructed one. 
\citet{Bengs_2021_AE_on_MRI} present an autoencoder-based method that also operates on voxel grids. A variational autoencoder is trained to reconstruct voxel grids through a low-dimensional latent variable. Anomaly scores are again derived by a per-voxel comparison of the input to its reconstruction.


\section{\uppercase{Description of the Dataset}}

\begin{table*}[!ht]
    \centering
    \resizebox{0.9\linewidth}{!}{%
    \begin{tabular}{cccccccc} 
    \hline
    Category     & \# Train & \begin{tabular}[c]{@{}c@{}}\# Val
    \end{tabular} & \begin{tabular}[c]{@{}c@{}}\# Test \\~(good)\end{tabular} & \begin{tabular}[c]{@{}c@{}}\# Test \\~(anomalous) \end{tabular} & \begin{tabular}[c]{@{}c@{}}\# Defect \\types\end{tabular} & \begin{tabular}[c]{@{}c@{}}\# Annotated \\regions\end{tabular} & \begin{tabular}[c]{@{}c@{}}Image size \\ (width $\times$ height) \end{tabular} \\
    \hline
    bagel       & 244      & 22     & 22         & 88       &  4    & 112   & 800 $\times$ 800  \\
    cable gland & 223      & 23     & 21         & 87       &  4    & 90    & 400 $\times$ 400  \\
    carrot      & 286      & 29     & 27         & 132      &  5    & 159   & 800 $\times$ 800  \\
    cookie      & 210      & 22     & 28         & 103      &  4    & 128   & 500 $\times$ 500  \\
    dowel       & 288      & 34     & 26         & 104      &  4    & 131   & 400 $\times$ 400  \\
    foam        & 236      & 27     & 20         & 80       &  4    & 115   & 900 $\times$ 900  \\
    peach       & 361      & 42     & 26         & 106      &  5    & 131   & 600 $\times$ 600  \\
    potato      & 300      & 33     & 22         & 92       &  4    & 115   & 800 $\times$ 800  \\
    rope        & 298      & 33     & 32         & 69       &  3    & 72    & 900 $\times$ 400  \\
    tire        & 210      & 29     & 25         & 87       &  4    & 95    & 600 $\times$ 800  \\
    
    \hline
    total       & 2656     & 294    & 249        & 948      & 41   & 1148   &       \\
    \hline
    \end{tabular}
    }
    \tabularspace
    \caption{Statistical overview of the MVTec 3D-AD dataset. For each category, we list the number of training, validation, and test images. Test images are split into anomaly-free images and images containing anomalies. We report the number of different defect types, the number of annotated regions, and the size of the $(x,y,z)$ images for each category.}
    \label{table:dataset_overview_table}
\end{table*}

The MVTec 3D-AD dataset consists of 4147 scans acquired by a high-resolution industrial 3D sensor. For each of the 10 object categories, a set of anomaly-free scans is provided for model training and validation. The test set contains both, anomaly-free scans as well as object samples that contain various types of anomalies, such as scratches, dents, or contaminations. The defects were devised and fabricated as they would occur in real-world inspection scenarios. 

Five of the object categories in our dataset exhibit considerable natural variations from sample to sample. These are \textit{bagel}, \textit{carrot}, \textit{cookie}, \textit{peach}, and \textit{potato}. Three more objects, \textit{foam}, \textit{rope}, and \textit{tire}, have a standardized appearance but can be easily deformed. The two remaining objects, \textit{cable gland} and \textit{dowel}, are rigid. In principle, inspecting the last two could be achieved by comparing an object's geometry to a CAD model. However, unsupervised methods should be able to detect anomalies on all kinds of objects and the creation of a CAD model might not always be desirable or practical in a real application. An example point cloud for each dataset category is shown in \Cref{fig:dataset_overview}. The figure also displays some anomalies together with the corresponding ground truth annotations. The images of the \textit{bagel} and the \textit{cookie} show cracks in the objects. The surfaces of the \textit{cable gland} and the \textit{dowel} exhibit geometrical deformations. There is a hole in the \textit{carrot} and some contaminations on the \textit{peach} and the \textit{rope}. Parts of the \textit{foam}, the \textit{potato}, and the \textit{tire} are cut off. These are prototypical examples of the 41 types of anomalies present in our dataset. More statistics on the dataset are listed in \Cref{table:dataset_overview_table}.

\begin{figure*}[!ht]
    \centering
    \includegraphics[width=0.98\textwidth]{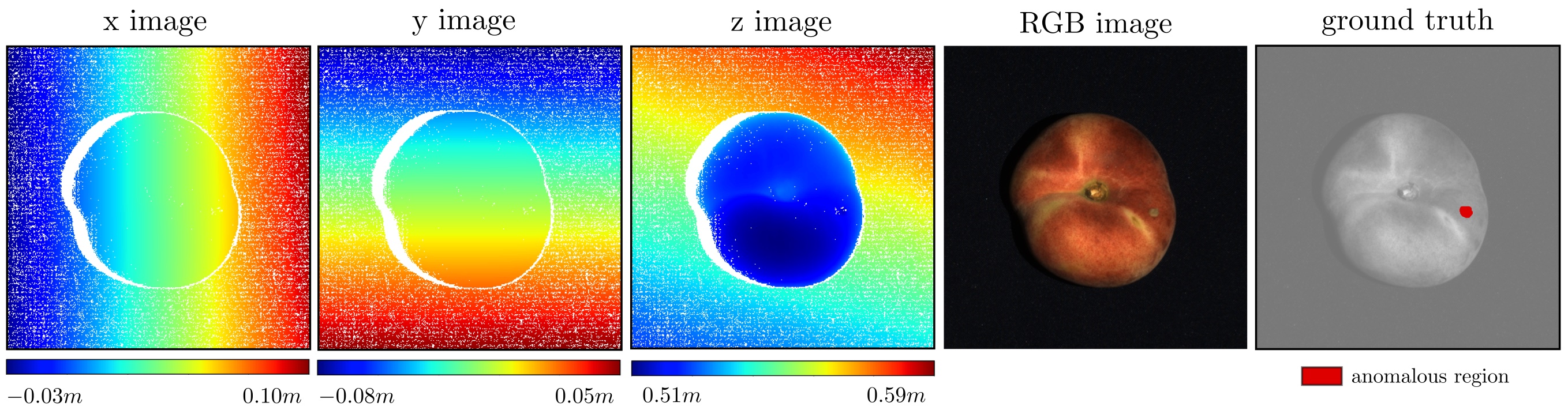}
    \caption{Visualization of the provided data for one anomalous test sample of the dataset category \textit{peach}. In addition to three images that encode the 3D coordinates of the object, RGB information as well as a pixel-precise ground-truth image are provided.}
    \label{fig:sensor_data}
\end{figure*}

\subsection{Data Acquisition and Preprocessing}
All dataset scans were acquired with a Zivid One$^+$ Medium,\footnote{\href{https://www.zivid.com/zivid-one-plus-medium-3d-camera}{https://www.zivid.com/zivid-one-plus-medium-3d-camera}} an industrial sensor that records high-resolution 3D scans using structured light. The data is provided by the sensor as a three-channel image with a resolution of 1920$\times$1200 pixels. The channels represent the $x$, $y$, and $z$ coordinates with respect to the local camera coordinate frame. The $(x,y,z)$ values of the image provide a one-to-one mapping to the corresponding point cloud. In addition, the sensor acquires complementary RGB values for each $(x,y,z)$ pixel. It was statically mounted to view all objects of each individual category from the same angle. We performed a calibration of the internal camera parameters that allows to project 3D points into the respective pixel coordinates \citep{steger-etal:18}. The scene was illuminated by an indirect and diffuse light source.

For each dataset category, we specified a fixed rectangular domain and cropped the original $(x,y,z)$ and RGB images to reduce the amount of background pixels in the samples. The acquisition setup as well as the preprocessing are very similar to real-world applications where an object is usually located in a defined position and the illumination is chosen to best suit the task. In addition, our setup enables and simplifies data augmentation. All objects were recorded on a dark background and the preprocessing leaves a sufficient margin around the objects to allow for the application of various data augmentation techniques, such as crops, translations, or rotations. This enables the use of our dataset for the training of data-hungry deep learning methods, as demonstrated by our experiments in \Cref{sec:benchmark}.

\Cref{fig:sensor_data} shows the data provided for the anomalous test sample of the \textit{peach} displayed in \Cref{fig:dataset_overview}. The image is of size 600$\times$600 pixels, cropped from the original sensor scan. The first three images visualize the $x$, $y$, and $z$ coordinates of the dataset sample, respectively. White pixels mark areas where the sensor did not return any 3D information due to, e.g., occlusions, reflections, or sensor inaccuracies. The corresponding RGB and ground truth annotation images are also displayed.

\subsection{Ground-Truth Annotations}
We provide precise ground-truth annotations for each anomalous sample in the test set. Anomalies were annotated in the 3D point clouds. Since there is a one-to-one mapping of the 3D points to their respective pixel locations in the $(x,y,z)$ image, we make the annotations available as two-dimensional regions. This procedure allows us to additionally label invalid sensor pixels and thus annotate anomalies that manifest themselves through the absence of points. For example, an anomaly might lead to a failure of 3D reconstruction and therefore yield invalid pixels in the 3D image. Furthermore, if an anomaly is visible in the RGB image and its corresponding color pixels are not already included in the ground truth label, we append these pixels to the annotation.

\begin{figure}[hb]
  \begin{center}
    \includegraphics[width=0.45\textwidth]{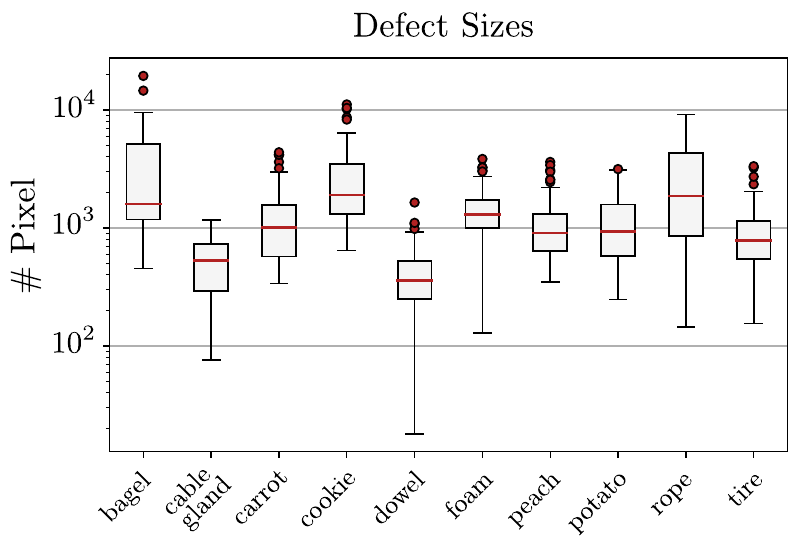}
  \end{center}
  \caption{Size of anomalies for all objects in the dataset visualized as a box-and-whisker plot. Defect areas are reported as the total number of pixels within an annotated connected component. Anomalies vary greatly in size for each dataset category.}
  \label{fig:defect_size_boxplot}
\end{figure}

An example ground truth mask is shown in \Cref{fig:sensor_data}, where a contamination is present on the \textit{peach}. In \Cref{fig:dataset_overview}, further annotations are visualized when projected to the valid 3D points of a scene. The size of the individual connected components of the anomalies present in the test set varies greatly, from a few hundred to several thousand pixels. \Cref{fig:defect_size_boxplot} visualizes their distribution as a box-and-whisker plot with outliers on a logarithmic scale \citep{tukey77}.

\subsection{Performance Evaluation}
\label{sec:performance_evaluation}

To assess the anomaly localization performance of a method on our dataset, we require it to output a real-valued anomaly score for each $(x,y,z)$ pixel of the test set. In contrast to only assigning anomaly scores to all valid 3D points of the test samples, this allows the detection of anomalies that manifest themselves through invalid sensor pixels or missing 3D structures. These anomaly scores are converted to binary predictions using a threshold.
We then compute the per-region overlap (PRO) metric \citep{Bergmann2021IJCV}, which is defined as the average relative overlap of the binary prediction $P$ with each connected component $C_{k}$ of the ground truth. The PRO value is computed as
\begin{equation}
    \mathrm{PRO} = \frac{1}{K} \sum_{k=1}^{K} \frac{|P \cap C_{k}|}{|C_{k}|},
\end{equation}
where $K$ is the total number of ground truth components. This process is repeated for multiple thresholds and a curve is constructed by plotting the resulting PRO values against the corresponding false positive rates. The final performance measure is computed by integrating this curve up to a limited false positive rate and normalizing the resulting area to the interval $[0,1]$. This is a standard metric for the task of unsupervised anomaly localization and is particularly useful when anomalies vary significantly in size.

We want to stress that, when working with our dataset, we strongly discourage to calculate the area under the PRO curve up to high false positive rates. We recommend to select the integration limit no larger than 0.3. This is due to the fact that the anomalous regions are very small compared to the size of the images. At large false positive rates, the amount of erroneously segmented pixels would be significantly larger than the number of actually anomalous pixels. This would lead to segmentation results that are no longer meaningful in practice. 

Our dataset can also be used to assess the performance of algorithms that make a binary decision for each sample if it contains an anomaly or not. In this case, we report the area under the ROC curve as a standard classification metric.\footnote{Evaluation scripts for the computation of all above performance measures will be published together with the dataset.}

\section{\uppercase{Initial Benchmark}}
\label{sec:benchmark}

To examine how existing 3D anomaly localization methods perform on our new dataset, we conducted an initial benchmark. It is intended to serve as a baseline for future methods. Very few methods have been proposed explicitly for this task and all of them operate on voxel data. This is mainly due to the fact that these methods are originally intended to process MR or CT scans that consist of several layers of intensity images. As representatives from this class of methods, we include Voxel f-AnoGAN \citep{Simarro_Viana_2021} and our own implementation of a convolutional Voxel AE \citep{Bengs_2021_AE_on_MRI} in our benchmark. Predecessors of these methods were developed for 2D image data. The main difference between the 2D and 3D methods is the use of 2D convolutions on images and 3D convolutions on voxel data, respectively. Therefore, these methods are easily adapted to process depth images and we include them in our benchmark as well. In addition to these deep learning methods, we evaluate the performance of variation models \citep{steger-etal:18} on voxel data and depth images. They detect anomalies by calculating the pixel- or voxel-wise Mahalanobis distance to the training data distribution. 

All evaluated methods can either operate solely on the 3D data or can additionally process the color information attached to each 3D point. We therefore also compare the difference in performance when adding color information to the models. Detailed information on training parameters and model architectures can be found in the Appendix.

\begin{table*}[t]
    \centering
    \resizebox{0.9\linewidth}{!}{%
    \begin{tabular}{ccc|cccccccccc|c} \hline
                              &                        &     & bagel & \begin{tabular}[c]{@{}c@{}}cable\\ gland\end{tabular} & carrot & cookie & dowel & foam & peach & potato & rope & tire & mean \\ \hline \hline
    \multirow{6}{*}{\rotatebox[origin=c]{90}{\textbf{3D Only}}}       & \multirow{3}{*}{\rotatebox[origin=c]{90}{Voxel}} & GAN & 0.440 & \textbf{0.453}  & \textbf{0.825} & \textbf{0.755} & \textbf{0.782} & \textbf{0.378} & \textbf{0.392} & 0.639 & \textbf{0.775} & \textbf{0.389} & \textbf{0.583} \\
                              &                        & AE & 0.260 & 0.341  & 0.581 & 0.351 & 0.502 & 0.234 & 0.351 & \textbf{0.658} & 0.015 & 0.185 & 0.348 \\
                              &                        & VM & \textbf{0.453} & 0.343  & 0.521 & 0.697 & 0.680 & 0.284 & 0.349 & 0.634 & 0.616 & 0.346 & 0.492 \\ \cline{2-14} 
                              & \multirow{3}{*}{\rotatebox[origin=c]{90}{Depth}} & GAN & 0.111 & 0.072  & 0.212 & 0.174 & 0.160 & 0.128 & 0.003 & 0.042 & 0.446 & 0.075 & 0.143 \\
                              &                        & AE & 0.147 & 0.069  & 0.293 & 0.217 & 0.207 & 0.181 & 0.164 & 0.066 & 0.545 & 0.142 & 0.203 \\
                              &                        & VM &  0.280 & 0.374  & 0.243 & 0.526 & 0.485 & 0.314 & 0.199 & 0.388 & 0.543 & 0.385 & 0.374 \\ \hline \hline
    \multirow{6}{*}{\rotatebox[origin=c]{90}{\textbf{3D + RGB}}} & \multirow{3}{*}{\rotatebox[origin=c]{90}{Voxel}}                 & GAN & \textbf{0.664} & 0.620  & 0.766 & \textbf{0.740} & 0.783 & 0.332 & 0.582 & 0.790 & 0.633 & 0.483 & \textbf{0.639} \\
                              &                        & AE  & 0.467 & \textbf{0.750}  & \textbf{0.808} & 0.550 & 0.765 & \textbf{0.473} & \textbf{0.721} & \textbf{0.918} & 0.019 & 0.170 & 0.564 \\
                              &                        & VM  & 0.510 & 0.331  & 0.413 & 0.715 & 0.680 & 0.279 & 0.300 & 0.507 & 0.611 & 0.366 & 0.471 \\ \cline{2-14} 
                              & \multirow{3}{*}{\rotatebox[origin=c]{90}{Depth}} & GAN & 0.421 & 0.422  & 0.778 & 0.696 & 0.494 & 0.252 & 0.285 & 0.362 & 0.402 & \textbf{0.631} & 0.474 \\
                              &                        & AE  & 0.432 & 0.158  & \textbf{0.808} & 0.491 & \textbf{0.841} & 0.406 & 0.262 & 0.216 & \textbf{0.716} & 0.478 & 0.481 \\
                              &                        & VM  & 0.388 & 0.321  & 0.194 & 0.570 & 0.408 & 0.282 & 0.244 & 0.349 & 0.268 & 0.331 & 0.335 \\ \hline
    \end{tabular}
    }
    \tabularspace
    \caption{Anomaly localization results. The area under the PRO curve is reported for an integration limit of $0.3$ for each evaluated method and dataset category. The best performing methods are highlighted in boldface.}
    \label{table:localization_results_quantitative}
\end{table*}

\subsection{Training and Evaluation Setup}

\paragraph{Data Representation.} To represent dataset samples as voxel grids, we first compute a global 3D bounding box over the entire training set for each dataset category. Then, a grid of $n \times n \times n$ voxels is placed at the center of the bounding box. The side length of the grid is is chosen to be equal to the longest side of the bounding box. If only 3D data is processed, occupied and empty voxels are assigned the values $1$ and $-1$, respectively. If RGB information is added, empty voxels are assigned the vector $(-1, -1, -1)$. Occupied voxels are assigned the average RGB value of all points that fall inside the same grid cell. 

For methods that process depth images, we compute the euclidean distance to the camera center for each valid pixel in the original $(x,y,z)$ images. Invalid pixels are assigned a distance of $0$. If color information is included, the RGB channels are appended to the single-channel depth image. For both, the voxel grids and depth images, the RGB values are scaled to the interval $[0, 1]$.

\paragraph{Methods on Voxel Grids.} For all voxel-based methods, we use grids of size $64 \times 64 \times 64$ voxels, as proposed by \citet{Simarro_Viana_2021}. To choose the latent dimension of the compression-based methods, we performed an ablation study, which is included in the Appendix. Anomaly scores are computed by a voxel-wise comparison of the input with its reconstruction.

The voxel grids of the samples of our dataset are sparsely populated and the majority of voxels is empty. We found that this leads to problems when training the Voxel AE if each voxel is weighted equally in the reconstruction loss. In this case, the model tends to simply output an empty voxel grid to minimize the reconstruction error. To address this imbalance, we introduce a loss weight $w \in (0, 1)$ that is computed as the fraction of empty voxels in the training set. During training, the loss at each voxel is then multiplied by $w$ if the voxel is occupied and by $(1-w)$ otherwise. 

For the Voxel Variation Model, we first compute the mean and standard deviation of the training data at each voxel. During inference, anomaly scores are derived by computing the voxel-wise Mahalanobis distance of each test sample to the training distribution.

\paragraph{Methods on Depth Images.} Our implementations of Depth f-AnoGAN and the Depth AE both process images at a resolution of $256 \times 256$ pixels. Input images are zoomed using nearest neighbor interpolation for depth, and bilinear interpolation for color images. Anomaly scores are derived by a per-pixel comparison of the input images and their reconstructions. The Depth Variation Model processes images with their original resolution and computes the mean and standard deviation over the entire training set at each image pixel. Again, anomaly scores are derived by computing the pixel-wise Mahalanobis distance from the training distribution.

\paragraph{Dataset Augmentation.} Since the evaluated methods, except for the Variation Models, require large amounts of training data, we use data augmentation to increase the size of the training set. For each object category, we first estimate the normal vector of the background plane, which is constant across samples. We then rotate each dataset sample around this normal vector by a certain angle and project the resulting points and corresponding color values into the original 2D image grid using the internal camera parameters. We augment each training sample 20 times by randomly sampling angles from the interval $[-5^\circ, 5^\circ]$.

\paragraph{Computation of Anomaly Maps.}
All voxel-based methods compute an anomaly score for each voxel element. However, comparing their performance on our dataset requires them to assign an anomaly score to each pixel in the original $(x,y,z)$ images. We therefore project the anomaly scores to pixel coordinates using the internal camera parameters of the 3D sensor. For each voxel element, we project all 8 corner points and compute the convex hull of the resulting projected points. All image pixels within this region are assigned the respective anomaly score of the voxel element. If a pixel is assigned multiple anomaly scores, we select their maximum. 

Methods on depth images already assign a score to each pixel. The anomaly maps of Depth f-AnoGAN and the Depth AE are zoomed to the original image size using bilinear interpolation.

\begin{figure*}[t]
    \centering
    \includegraphics[width=0.83\textwidth]{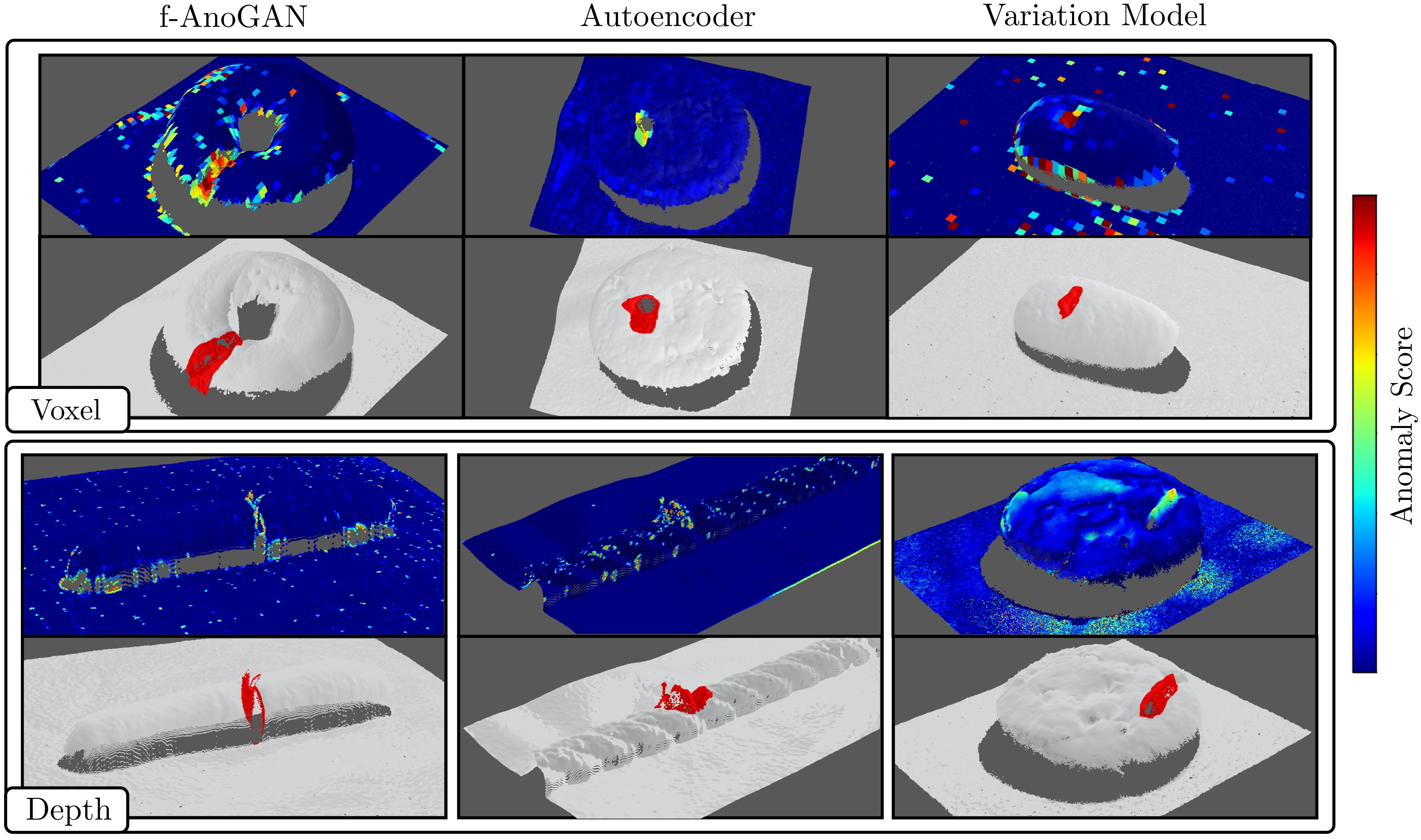}
    \caption{Qualitative anomaly localization results in which each individual method performed well. The top image visualizes the anomaly scores as an overlay to the input point cloud. The bottom image shows the corresponding ground-truth annotation in red. The displayed methods were only trained on the 3D data without adding color information.}
    \label{fig:qualitative_results}
\end{figure*}

\subsection{Results}

\Cref{table:localization_results_quantitative} lists quantitative results for each evaluated method for the localization of anomalies. For each dataset category, we report the normalized area under the PRO curve with an upper integration limit of $0.3$. We further report the mean performance over all categories. Here, we focus on the localization performance of the evaluated methods. Results on sample level anomaly classification can be found in the Appendix.

The first six rows in \Cref{table:localization_results_quantitative} show the performance of each method when trained only on the 3D sensor data without providing any color information. In this case, the Voxel f-AnoGAN performs best on average and on the majority of all dataset categories. It is followed by the Voxel VM, which shows the best performance on one of the objects. The Voxel AE performs worse than the other two voxel-based methods. This is because it tends to produce blurry and inaccurate reconstructions. To get an impression of the reconstruction quality of the evaluated methods, we show some qualitative examples in the Appendix.

On average, each voxel-based method performs better than its depth-based counterpart. Among all depth-based methods, the Depth Variation Model performs best. We found that the Depth AE and Depth f-AnoGAN produce many false positives in the anomaly maps around invalid pixels in the input data.

\Cref{fig:qualitative_results} shows corresponding qualitative anomaly localization results. For visualization purposes, the predicted anomaly scores were projected onto the input point clouds. For each dataset sample, the corresponding ground truth is visualized in red. While most of the methods are able to localize some of the defects in our dataset, they also yield a large number of false positive predictions, either on the objects' surfaces, around the objects' edges, or in the background. Due to the reconstruction inaccuracies of the Voxel AE, it can only detect the larger and more salient anomalies in our dataset such as the one depicted in \Cref{fig:qualitative_results}.

\begin{figure}[b]
  \begin{center}
    \includegraphics[width=0.4\textwidth]{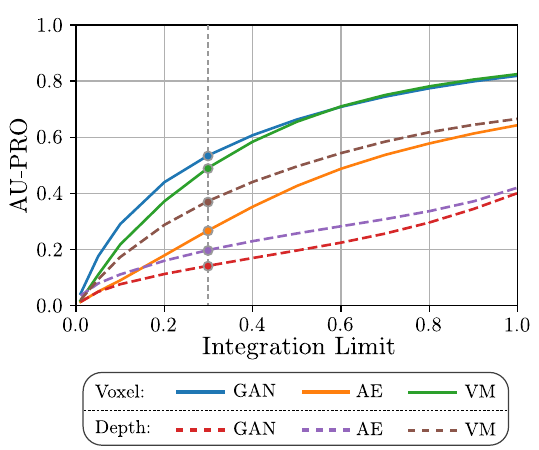}
  \end{center}
  \caption{Dependence of AU-PRO on the integration limit. The results of our benchmark are reported at a limit of $0.3$.}
  \label{fig:ablation_integration_limit}
\end{figure}

In addition to evaluating each method on 3D data only, we report the performance of the methods when trained with RGB features at each 3D point. The results are listed in the bottom six rows of \Cref{table:localization_results_quantitative}.
Adding RGB information improves the performance of all methods except for the Variation Models. Since the RGB images do not contain invalid pixels, the Depth AE and Depth f-AnoGAN benefit most from the color information. Nevertheless, the voxel-based methods still outperform their depth-based counterparts. Again, Voxel f-AnoGAN shows the best overall performance. For some object categories, however, the Voxel AE performs better than Voxel f-AnoGAN when including color information.

As discussed in \Cref{sec:performance_evaluation}, it is important to select a suitable integration limit to compute the area under the PRO curve. To illustrate this, \Cref{fig:ablation_integration_limit} shows the dependence of the performance of each evaluated method on the integration limit.
All methods show a monotonic increase in their AU-PRO. When integrating to a false positive rate of $1$, the Voxel f-AnoGAN and the Voxel VM surpass an AU-PRO of $0.8$, which would suggest that these methods are close to solving the task. However, evaluating at large integration limits include binary segmentation masks where the number of false positive predictions is extremely high. Since the area of the defects present in our dataset is very small compared to the area of anomaly-free pixels, such segmentation results are no longer meaningful. We found that the performance of the evaluated methods is insufficient for practical applications. We therefore chose an integration limit of $0.3$ to be able to compare their relative performance. In the future, we hope that our dataset sparks the development of methods that achieve high AU-PRO scores at much lower integration limits. In actual industrial inspection scenarios, a false positive rate of $0.3$ is hardly acceptable.

\section{\uppercase{Conclusion}}
We presented a comprehensive 3D dataset for the task of unsupervised detection and localization of anomalies. The conceptualization and acquisition of the dataset was inspired by real-world visual inspection tasks. It consists of over 4000 point clouds depicting instances of ten different object categories. The data was acquired using a high-resolution structured light 3D sensor. About 1000 samples of the dataset contain various types of anomalies and we provide precise ground truth annotations for all of them.

We performed an initial benchmark of the few existing methods showing that there is significant room for improvement. In particular, the accuracy of the evaluated methods is insufficient for them to be used in real-world industrial applications. We are convinced that suitable datasets are a key factor in the development of new techniques and expect our dataset to spark the design of better methods in the future. 

\bibliographystyle{apalike}
{\small
\bibliography{references}}

\section*{\uppercase{Appendix}}

\subsection*{Details on Training Parameters.}
\label{sec:additional_training_details}

In this section, we provide details on training parameters as well as model architectures for the deep learning-based methods.

\paragraph{Voxel f-AnoGAN}
For the implementation of Voxel f-AnoGAN, we use the same network architecture as proposed by \citet{Simarro_Viana_2021}. 
The GAN and the encoder network are both trained for $50$ epochs on the augmented version of our dataset with an initial learning rate of $0.0002$ and a batch size of $2$ using the Adam optimizer \citep{kingma2015adam}. The weight for the gradient penalty loss of the GAN is set to $10$ and one generator training iteration is performed for every $5$ iterations of the discriminator training. During the training of the encoder, the \enquote{izi} and \enquote{ziz} losses are equally weighted by choosing a loss weight of $1$. 
 
\paragraph{Voxel Autoencoder}
The Voxel Autoencoder consists of an encoder and a decoder network. Their architecture is the same as the one of the encoder and the generator in Voxel f-AnoGAN, respectively. We train for $50$ epochs on the augmented version of our dataset with a batch size of $2$ using the Adam optimizer with an initial learning rate of $0.0001$.

\paragraph{Depth f-AnoGAN}
The Depth f-AnoGAN consists of three sub-networks, i.e., an encoder, a discriminator, and a generator. The architecture of the encoder is given in \Cref{table:encoder_architecture_2d}. It consists of a stack of $10$ convolution blocks that compress an input image of size $256 \times 256$ pixels and $c$ channels to a $d$-dimensional latent vector. Each convolution block except the last one is followed by an instance normalization layer \citep{Ulyanov_2017_CVPR} and a LeakyReLU with slope $0.05$. The architecture of the discriminator is identical to the one of the encoder except that $d=1$. The generator produces an image of size $256 \times 256$ pixels and $c$ channels from a latent variable with $d$ dimensions. Its architecture is symmetric to the one in \Cref{table:encoder_architecture_2d} in the sense that convolutions are replaced by transposed convolution layers.
Both, the GAN and the encoder network, are trained for $50$ epochs using a batch size of $4$ and an initial learning rate of $0.0002$ using the Adam optimizer. During the training of the encoder, the \enquote{izi} and \enquote{ziz} losses are equally weighted by choosing a loss weight of $1$. 

\begin{table}[hb]
    \centering
    \resizebox{\linewidth}{!}{%
    \begin{tabular}{lrccc}
    \hline
    \multicolumn{1}{c}{Layer} & \multicolumn{1}{c}{Output Size} & \multicolumn{3}{c}{Parameters} \\
    \multicolumn{1}{c}{} & \multicolumn{1}{c}{} & \multicolumn{1}{c}{Kernel Size} & \multicolumn{1}{c}{Stride} & \multicolumn{1}{c}{Padding} \\ \hline \hline
    \multicolumn{1}{l}{Image} & \multicolumn{1}{r}{256 $\times$ 256 $\times$ $\phantom{12}c$} & \multicolumn{1}{c}{} & \multicolumn{1}{c}{} & \multicolumn{1}{c}{} \\
    conv1 & 128 $\times$ 128 $\times\phantom{1}$ 32 & 4 & 2 & 1 \\
    conv2 & 64 $\times\phantom{1}$ 64 $\times\phantom{1}$ 32 & 4 & 2 & 1 \\
    conv3 & 32 $\times\phantom{1}$ 32 $\times\phantom{1}$ 32 & 4 & 2 & 1 \\
    conv4 & 32 $\times\phantom{1}$ 32 $\times\phantom{1}$ 32 & 3 & 1 & 1 \\
    conv5 & 16 $\times\phantom{2}$ 16 $\times\phantom{1}$ 64 & 4 & 2 & 1 \\
    conv6 & 16 $\times\phantom{2}$ 16 $\times\phantom{1}$ 64 & 3 & 1 & 1 \\
    conv7 & 8 $\times\phantom{12}$ 8 $\times$ 128 & 4 & 2 & 1 \\
    conv8 & 8 $\times\phantom{12}$ 8 $\times\phantom{1}$ 64 & 3 & 1 & 1 \\
    conv9 & 8 $\times\phantom{12}$ 8 $\times\phantom{1}$ 32 & 3 & 1 & 1 \\
    conv10 & 1 $\times\phantom{28}$ 1 $\times\phantom{12}$ $d$ & 8 & 1 & 0 \\ \hline
    \end{tabular}
    }
    \tabularspace
    \caption{Encoder architecture of Depth f-AnoGAN and Depth AE.}
    \label{table:encoder_architecture_2d}
\end{table}

\paragraph{Depth Autoencoder}
For the encoder and decoder of the Depth AE, we use the same architecture as for the encoder and generator of Depth f-AnoGAN, respectively. It is trained for $50$ epochs using the Adam optimizer at a batch size of $32$ and an initial learning rate of $0.0001$.

\subsection*{Latent Dimensions of Compression-Based Methods.}
\label{sec:ablation_latent}
To select suitable latent dimensions for the evaluated compression-based methods, we perform an ablation study. Their mean performance over all object categories is given in \Cref{fig:ablation_latent}. For the experiments in \Cref{sec:benchmark}, we use the respective latent dimension that yielded the best mean performance in the ablation study.

\begin{table}[ht]
\centering
{%
\begin{tabular}{cc|ccc}
\hline
 &  & \multicolumn{3}{c}{Latent Dimension} \\
 &  & 128 & 512 & 2048 \\ \hline \hline
\multirow{2}{*}{Voxel} & GAN & 0.536 & \textbf{0.583} & 0.555 \\
 & AE & \textbf{0.348} & 0.269 & 0.305 \\ \hline
\multirow{2}{*}{Depth} & GAN & \textbf{0.143} & 0.137 & 0.135 \\
 & AE & 0.199 & \textbf{0.203} & 0.199 \\ \hline
\end{tabular}
}
\tabularspace
\caption{Difference in performance when varying the latent dimension of each compression-based method. We list the area under the PRO curve up to an integration limit of $0.3$. The best performing setting is highlighted in boldface.}
\label{fig:ablation_latent}
\end{table}

\subsection*{Results for Anomaly Classification.}
\label{sec:classification_results}

In addition to the anomaly localization results in \Cref{table:localization_results_quantitative}, we provide results for the classification of dataset samples as either anomalous or anomaly-free. Since this requires a method to output a single anomaly score for each dataset sample, we compute the maximum anomaly score of each anomaly map. As performance measure, we compute the area under the ROC curve. \Cref{table:classification_results_quantitative} lists the results. 

\begin{table*}[ht]
\centering
\resizebox{0.9\linewidth}{!}{%
\begin{tabular}{ccc|cccccccccc|c} \hline
                          &                        &     & bagel & \begin{tabular}[c]{@{}c@{}}cable\\ gland\end{tabular} & carrot & cookie & dowel & foam & peach & potato & rope & tire & mean \\ \hline \hline
\multirow{6}{*}{\rotatebox[origin=c]{90}{\textbf{3D Only}}}       & \multirow{3}{*}{\rotatebox[origin=c]{90}{Voxel}} & GAN & 0.383 & 0.623  & 0.474 & 0.639 & 0.564 & 0.409 & 0.617 & 0.427 & 0.663 & 0.577 & 0.538 \\
                          &                        & AE & 0.693 & 0.425  & 0.515 & \textbf{0.790} & 0.494 & 0.558 & 0.537 & 0.484 & 0.639 & 0.583 & 0.572 \\
                          &                        & VM & \textbf{0.750} & \textbf{0.747}  & \textbf{0.613} & 0.738 & \textbf{0.823} & 0.693 & \textbf{0.679} & \textbf{0.652} & 0.609 & \textbf{0.690} & \textbf{0.699} \\ \cline{2-14} 
                          & \multirow{3}{*}{\rotatebox[origin=c]{90}{Depth}} & GAN & 0.530 & 0.376  & 0.607 & 0.603 & 0.497 & 0.484 & 0.595 & 0.489 & 0.536 & 0.521 & 0.524 \\
                          &                        & AE & 0.468 & 0.731  & 0.497 & 0.673 & 0.534 & 0.417 & 0.485 & 0.549 & 0.564 & 0.546 & 0.546 \\
                          &                        & VM &  0.510 & 0.542  & 0.469 & 0.576 & 0.609 & \textbf{0.699} & 0.450 & 0.419 & \textbf{0.668} & 0.520 & 0.546 \\ \hline \hline
\multirow{6}{*}{\rotatebox[origin=c]{90}{\textbf{3D + RGB}}} & \multirow{3}{*}{\rotatebox[origin=c]{90}{Voxel}}                 & GAN & \textbf{0.680} & 0.324  & 0.565 & 0.399 & 0.497 & 0.482 & 0.566 & 0.579 & 0.601 & 0.482 & 0.517 \\
                          &                        & AE  & 0.510 & 0.540  & 0.384 & 0.693 & 0.446 & 0.632 & 0.550 & 0.494 & \textbf{0.721} & 0.413 & 0.538 \\
                          &                        & VM  & 0.553 & \textbf{0.772}  & 0.484 & \textbf{0.701} & 0.751 & 0.578 & 0.480 & 0.466 & 0.689 & 0.611 & \textbf{0.609} \\ \cline{2-14} 
                          & \multirow{3}{*}{\rotatebox[origin=c]{90}{Depth}} & GAN & 0.538 & 0.372  & 0.580 & 0.603 & 0.430 & 0.534 & 0.642 & \textbf{0.601} & 0.443 & 0.577 & 0.532 \\
                          &                        & AE  & 0.648 & 0.502  & \textbf{0.650} & 0.488 & \textbf{0.805} & 0.522 & \textbf{0.712} & 0.529 & 0.540 & 0.552 & 0.595 \\
                          &                        & VM  & 0.513 & 0.551 & 0.477 & 0.581 & 0.617 & \textbf{0.716} & 0.450 & 0.421 & 0.598 & \textbf{0.623} & 0.555 \\ \hline
\end{tabular}
}
\tabularspace
\caption{Anomaly classification results. We report the area under the ROC curve. The best-performing methods are highlighted in boldface.}
\label{table:classification_results_quantitative}
\end{table*}

\subsection*{Quality of Reconstructions.}

\begin{figure*}[ht]
    \centering
    \includegraphics[width=0.9\textwidth]{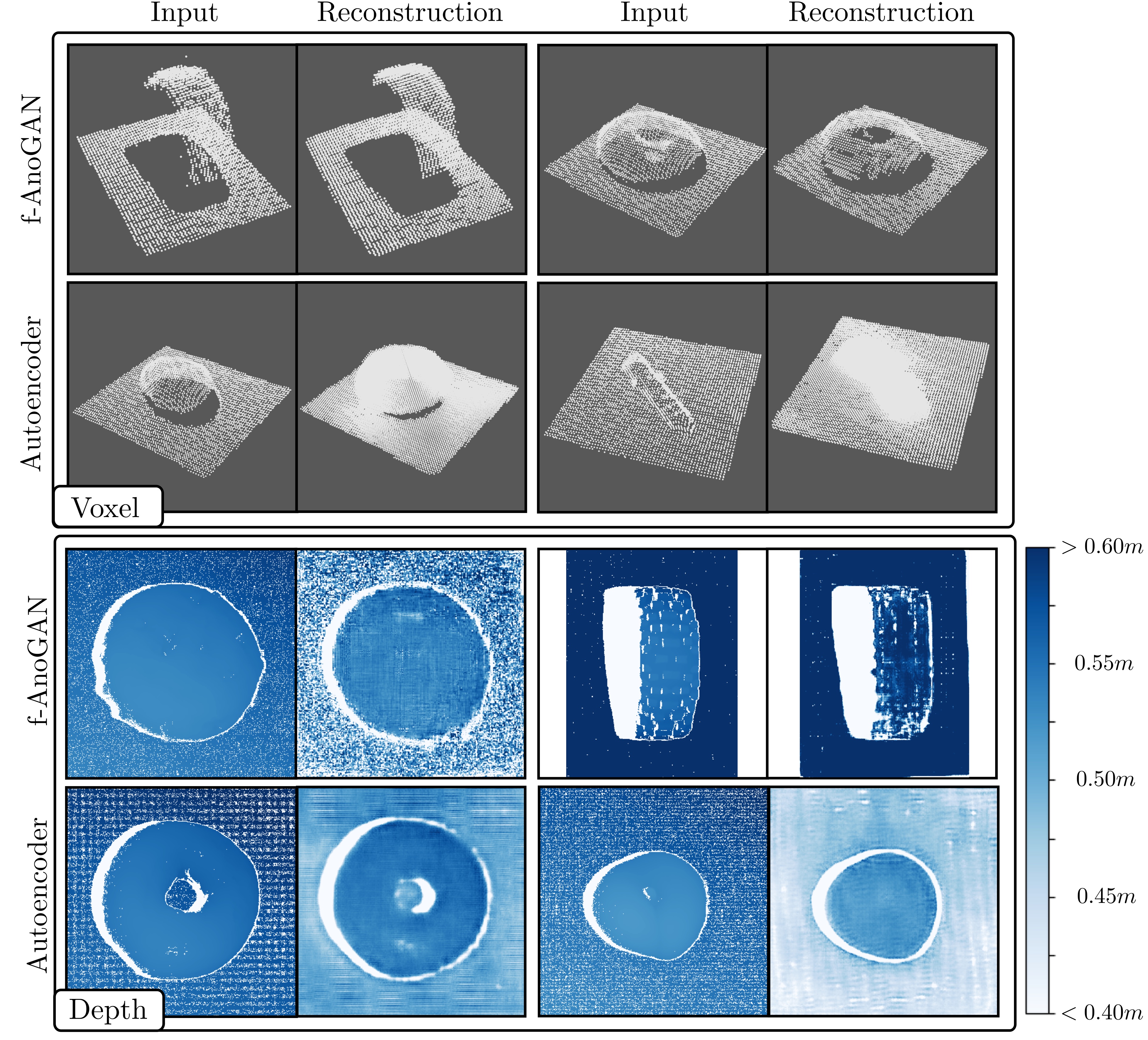}
    \caption{Examples of reconstructions for each compression-based method. For voxel-based methods, inputs and reconstructions are visualized as point clouds. For depth-based methods, they are shown as depth images.}
    \label{fig:qualitative_reconstructions}
\end{figure*}

For the methods based on AEs or GANs, the anomaly detection performance significantly depends on the quality of their reconstructions. To get an impression of the reconstruction quality, \Cref{fig:qualitative_reconstructions} shows two examples for each evaluated method. 

To visualize the voxel-based methods, voxel grids are converted to point clouds by applying a threshold to each cell. A cell is classified as occupied if it contains a value of 0.9 or higher. The Voxel AE tends to produce blurry reconstructions around the objects' surfaces. The Voxel f-AnoGAN does not have this problem. However, it sometimes fails to produce parts of the input.

For the depth-based methods, inputs and reconstructions are visualized as depth images. Darker shades of blue indicate points that are further from the camera center. White points indicate invalid pixels. Both, the Depth f-AnoGAN and the Depth AE show problems reconstructing noisy areas that exhibit many invalid pixels. 

\label{sec:qualitative_reconstructions}

\end{document}